\documentclass{article}

\PassOptionsToPackage{numbers, compress}{natbib}

\usepackage[eandd, preprint]{neurips_2026}

\usepackage{microtype}
\usepackage{graphicx}
\usepackage{subcaption}
\usepackage{booktabs} 
\usepackage{svg}
\usepackage{float}
\usepackage[dvipsnames]{xcolor}
\usepackage[pagebackref=true,colorlinks,citecolor=RoyalBlue,bookmarks=false]{hyperref}
\usepackage{colortbl}
\usepackage[normalem]{ulem}
\usepackage{enumitem}

\usepackage{amsmath}
\usepackage{amssymb}
\usepackage{mathtools}
\usepackage{amsthm}
\usepackage{multirow}
\usepackage{xspace}
\usepackage{stfloats}
\usepackage{tikz}
\usepackage{svg}
\usepackage[most]{tcolorbox}

\usepackage[utf8]{inputenc} 
\usepackage[T1]{fontenc}    
\usepackage{url}            
\usepackage{amsfonts}       
\usepackage{nicefrac}       
\usepackage{wrapfig}
\usepackage{float}

\usepackage{subcaption}  

\usepackage[capitalize,noabbrev]{cleveref}

\newcommand{\bmname}{VidMsg\xspace}
\newcommand{\ours}{VidVec-Msg\xspace}

\title{VidMsg: A Benchmark for Implicit Message Inference in Short Videos}

\author{%
  Issar Tzachor
  \And
  Michael Green \\
  \\
  OriginAI, Israel \\
  \vspace{3mm}
  \\
  \textcolor[HTML]{5BC9C5}{\nolinkurl{https://iyttor.github.io/VidMsg}} \\
  \vspace{-6.5mm}
  \And
  Rami Ben-Ari
}

\begin{document}

\maketitle

\begin{abstract}
Understanding short online videos involves more than identifying visible objects and actions; video makers often include an underlying message or purpose in the clip. We introduce VidMsg, a benchmark for evaluating implicit message understanding in short, internet-native video clips. VidMsg contains 400 YouTube-derived clips across 9 practical topic areas and 52 fine-grained target messages, covering domains such as career and finance, education, health and well-being, culture, safety, sustainability, and lifestyle.
VidMsg is constructed through a message-first pipeline: an LLM first translates target messages into indirect search scenarios, which are used to retrieve candidate clips. Human annotators then retain clips that convey the intended message without being overly explicit.
VidMsg is designed primarily for bidirectional message–clip retrieval for scalable applications such as video search and recommendation, where systems must capture holistic video understanding.
In addition to retrieval, VidMsg includes a diagnostic multiple-choice QA benchmark, where models select the intended message of a clip from semantically related alternatives.
Experiments with contemporary video-language and retrieval models show that strong models often fail on VidMsg, because the task requires pragmatic inference, integration of contextual cues, and discrimination among semantically close messages. We also introduce VidVec-Msg, a baseline method that improves message-oriented retrieval while leaving substantial headroom for future work.
\end{abstract}

\begin{figure}[H]
    \centering
    \begin{minipage}[b]{0.42\linewidth}
    \includegraphics[width=\linewidth]{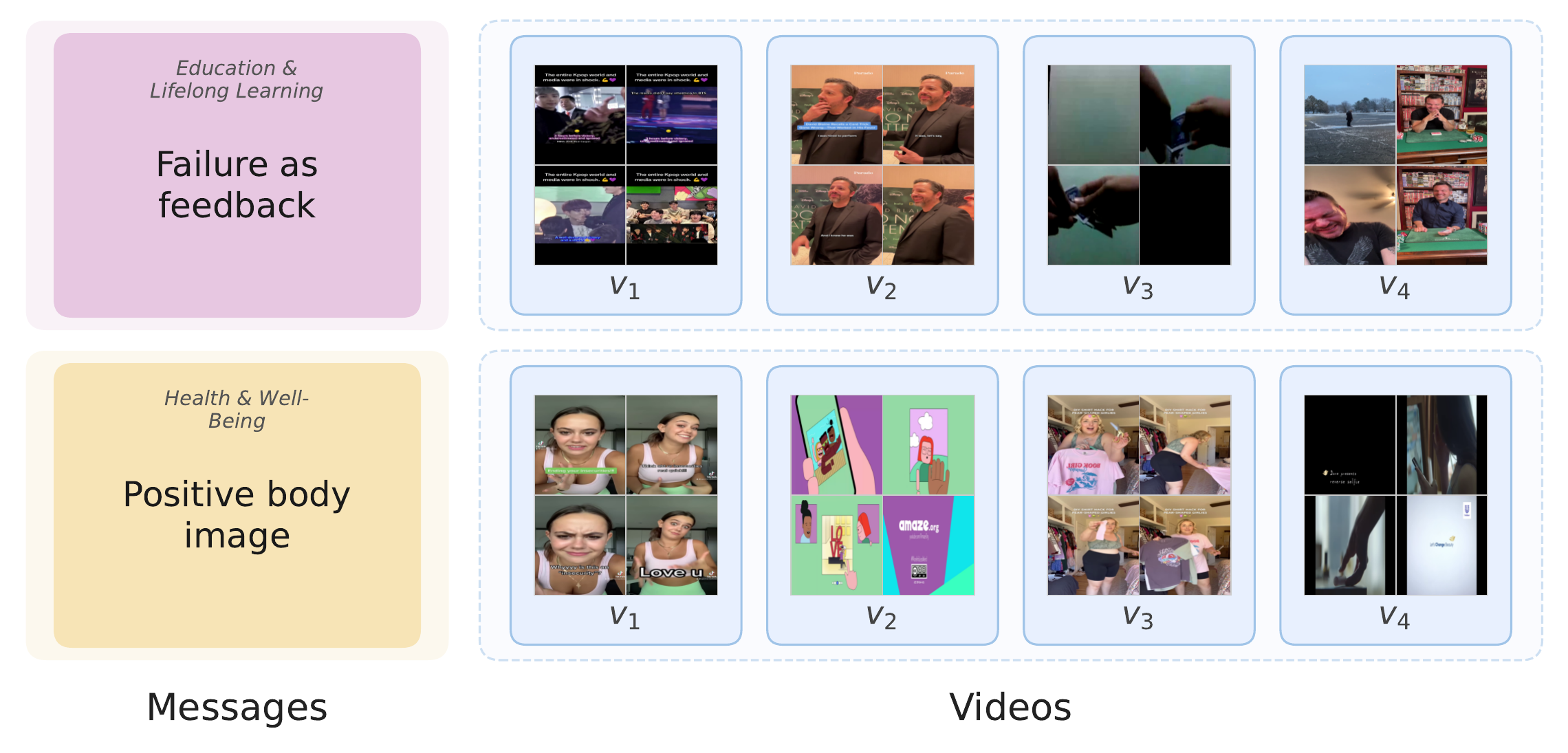}
    \label{fig:Schematic_Examples}
    
    \includegraphics[width=\linewidth]{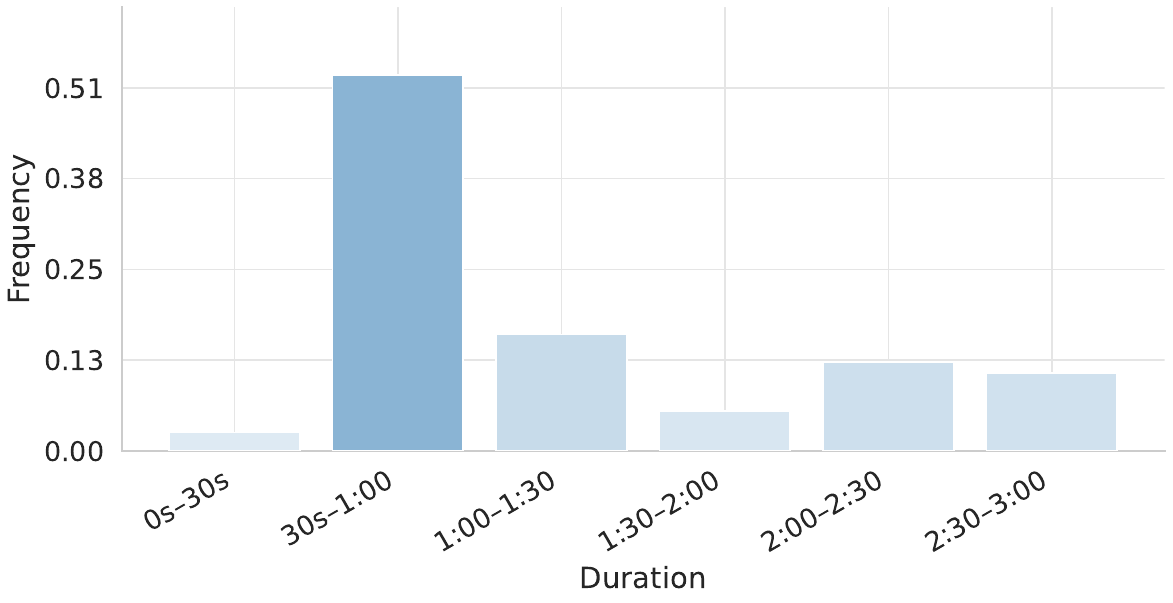}
    \captionsetup{font=small}
    \caption{\bmname{}: Examples \& Clip Durations}
    \label{fig:Schematic Example and Clip Duration}
    \end{minipage}
    \hfil
    \begin{minipage}[b]{0.45\linewidth}
    \includegraphics[width=\linewidth]{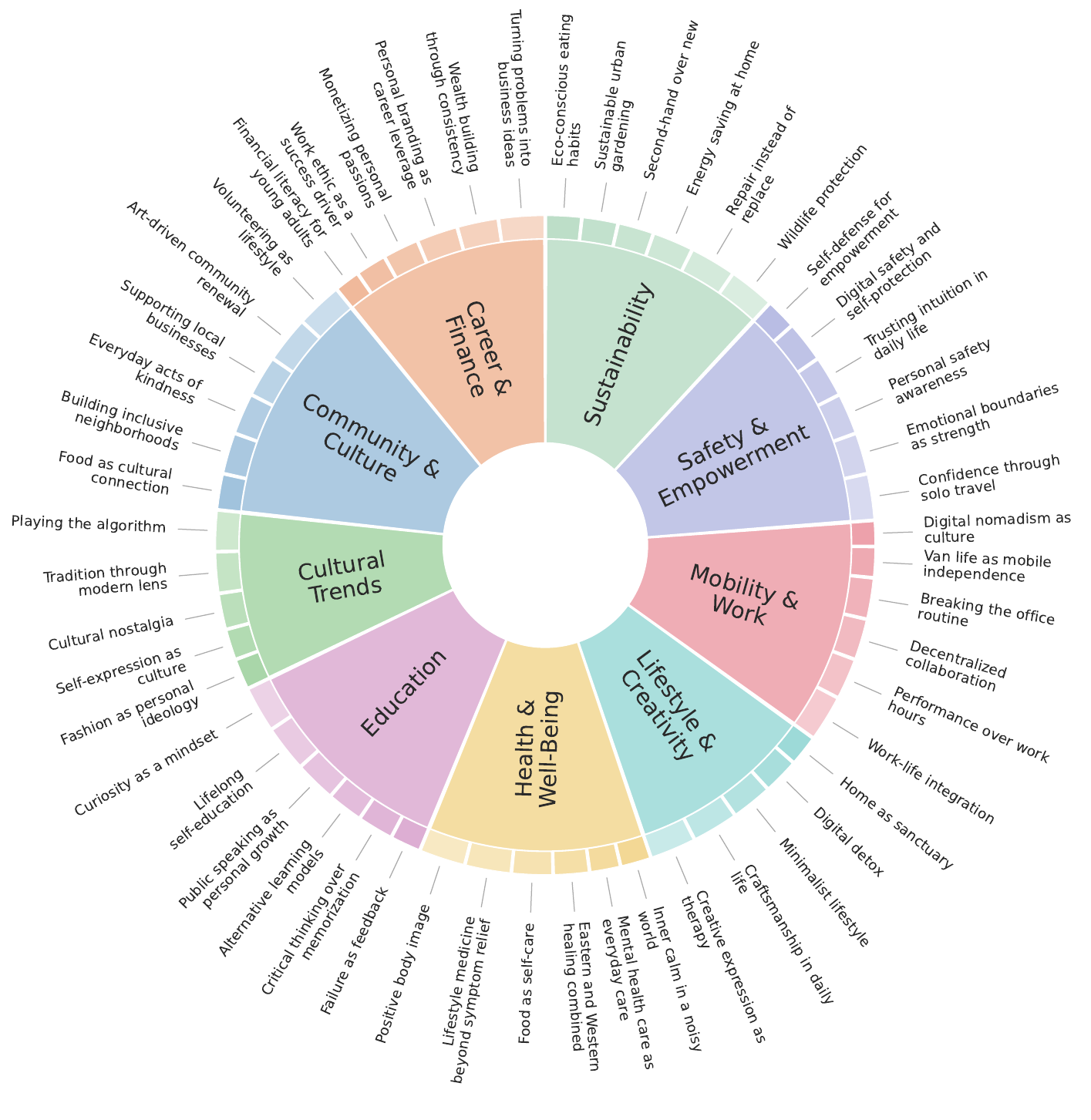}
    \captionsetup{font=small}
    \caption{VidMsg: Topics and Messages}
    \label{fig:VidMsg Topic Distribution}
    \end{minipage}
    \vspace{-3mm}
\end{figure}

\section{Introduction}
\label{sec:intro}
Understanding a video often requires looking beyond the visible objects and actions. Viewers also infer the broader message a clip is trying to communicate: the high-level proposition, attitude, or interpretation that the video encourages them to adopt. This message is distinct from a caption, which describes what is shown or said; from an actor's intent, which concerns what a person in the scene wants; and from a topic, which provides only a coarse semantic category. A video message is instead the communicative meaning implied by the clip as a whole.

This distinction is especially important for short, internet-native videos. Such clips are often consumed, searched, shared, and recommended because of the message they convey rather than their literal visual content. A user may look for videos about perseverance, body acceptance, environmental responsibility, or distrust of consumer culture even when those phrases are never spoken or shown. The relevant evidence may be distributed across visual cues, speech, text overlays, facial expressions, narrative contrast, and cultural context. For example, a clip showing a person struggling with ill-fitting clothes may communicate that discomfort with clothing should not be interpreted as a failure of one's body, but as a mismatch between diverse bodies and standardized clothing design. Inferring this message requires pragmatic and multimodal reasoning beyond object recognition, action recognition, or surface-level captioning.

Video-language benchmarks have driven substantial progress in multimodal understanding. Early and foundational datasets evaluate capabilities such as action recognition, temporal localization, video captioning, and video-text retrieval~\cite{kay2017kinetics,goyal2017something,yu2019activitynetqa,hendricks2017didemo,xu2016msrvtt}. More recent benchmarks extend evaluation to causal and temporal reasoning, long-form video understanding, and multimodal large language model (MLLM) assessment~\cite{xiao2021nextqa,li2023intentqa,fu2025videomme,mangalam2023egoschema,wang2024lvbench,zhou2025mlvu}. These benchmarks are complementary to our goal; they primarily ask what is visible, what happened, when it happened, or why one observable event followed another. Even benchmarks targeting higher-level reasoning, such as intent-sensitive~\cite{xiao2021nextqa, li2023intentqa} or advertisement-focused evaluation~\cite{long2025adsqa}, do not directly isolate the general problem of inferring the implicit communicative message of everyday short video clips.

We introduce \bmname{}, a benchmark for implicit message understanding in real-world short video clips. \bmname{} evaluates whether models can identify and retrieve videos according to what they communicate, rather than only according to what they depict. The benchmark contains 400 curated clips spanning 9 practical topic areas and 52 fine-grained target messages, including career and finance, education, health and well-being, culture, safety, sustainability, and lifestyle. These topics were selected because short videos in such domains often convey abstract advice, values, emotions, or social meanings that are not directly stated.

\bmname{} is designed primarily around a bidirectional message--clip retrieval setting. In a bidirectional retrieval task, a model must retrieve clips given a message query or retrieve the intended message given a clip. This setting is important for scalable applications such as search, recommendation, clustering, content analysis, and educational media discovery, where systems must represent the holistic meaning of a video in a compact embedding rather than rely only on visible entities or events. In addition to retrieval, \bmname{} includes a multiple-choice question-answering (MCQ) task as a diagnostic analysis tool: given a clip, a model selects the intended message from semantically related alternatives within the same topic. This MCQ setting helps probe whether models can discriminate intended messages, under a common QA evaluation protocol.

Constructing such a benchmark is challenging because direct web search often returns clips where the target message is stated explicitly in the title, speech, or text overlay. To address this, we use a message-first data collection pipeline. Starting from a target message, we generate narrative scenarios that imply the message without stating it directly. These scenarios are used to produce indirect search queries for retrieving candidate real-world clips. Candidate clips are then annotated by multiple human workers, who rate how strongly each clip conveys the target message and flag cases where the message is too explicit. We retain only clips with majority high-relevance judgments, no very-low relevance judgment, and no ``too obvious'' flag. This conservative filtering procedure yields a benchmark focused on relevant but non-trivial examples of implicit message understanding.

Our experiments show that strong contemporary video-language and retrieval models remain far from reliable on \bmname{}. Models that perform well on conventional video benchmarks often fail here because the task requires pragmatic inference, integration of weak multimodal cues, and discrimination among semantically close messages.
In many cases, models may describe the surface content of a clip correctly but still select a generic or incorrect message, suggesting a gap between perceptual recognition and faithful message understanding.

Finally, we introduce \ours{}, a baseline method for message-oriented retrieval. Rather than treating implicit message understanding as standard captioning or video QA, \ours{} builds on \cite{vidvec2026} through lightweight text-only adaptation. Specifically, we extend their approach to synthetic dense clip-storyline--message pairs, encouraging the model to identify visual and contextual cues associated with implied messages. This improves retrieval performance on \bmname{}, while leaving substantial headroom for future work.

Our contributions are threefold:
(i) We introduce \bmname{}, a benchmark for implicit message understanding in short, real-world, internet-native video clips.
(ii) We target message-level video retrieval and define complementary multiple-choice and bidirectional evaluation protocols for assessing message-level comprehension.
(iii) We provide a message-first data construction pipeline with human annotations of relevance and explicitness, conduct extensive model evaluations, and introduce a baseline method for message-oriented retrieval, evaluated both with and without video transcripts as inputs to MLLMs.

\begin{figure}[t]
\vspace{-7mm}
    \centering

    \begin{subfigure}[t]{0.47\linewidth}
        \centering
        \includegraphics[width=\linewidth]{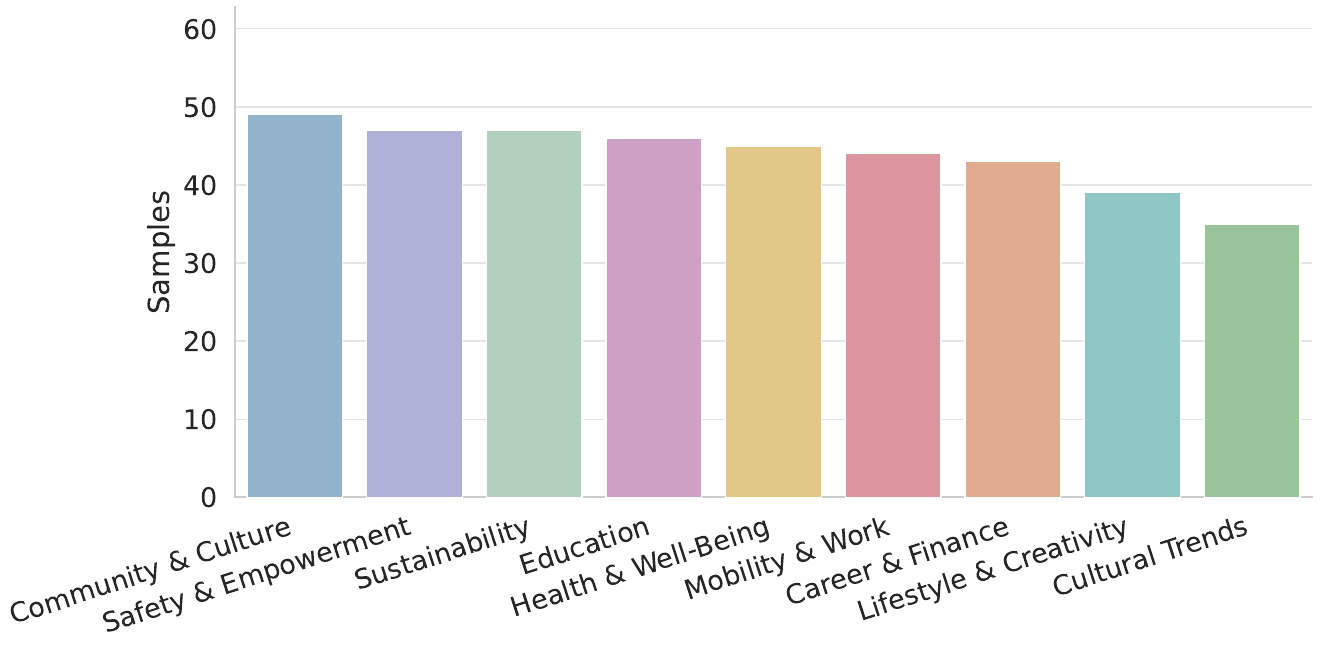}
        \captionsetup{font=small}
        \caption{Total number of clips per topic.}
        \label{fig:vids_per_topic}
    \end{subfigure}
    \hfill
    \begin{subfigure}[t]{0.47\linewidth}
        \centering
        \includegraphics[width=\linewidth]{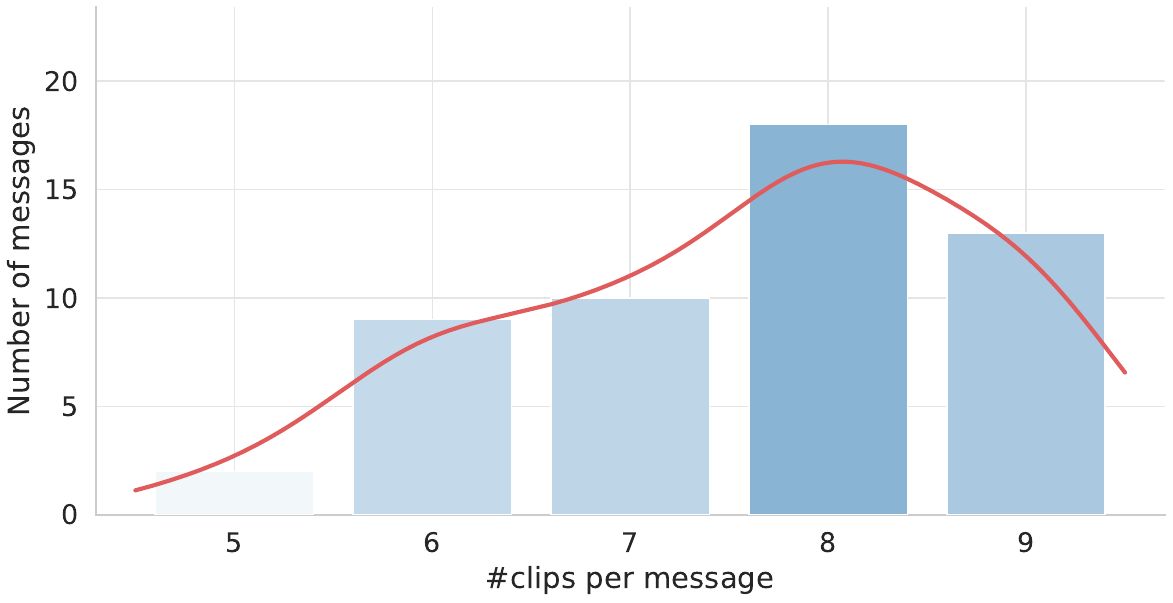}
        \captionsetup{font=small}
        \caption{Distribution of the number of clips per message.}
        \label{fig:vids_per_msg}
    \end{subfigure}
    \caption{Distribution of videos across topics and messages.}
    \label{fig:videos_distribution}
\vspace{-4mm}
\end{figure}

\section{Related Work}
\label{sec:related}
\paragraph{Action- and event-centric video benchmarks.}
Early video benchmarks primarily drove progress in visual recognition and event understanding. Datasets such as Kinetics~\cite{kay2017kinetics}, UCF101~\cite{soomro2012ucf101}, and HMDB51~\cite{kuehne2011hmdb} evaluate whether models can recognize actions or activities from visual motion patterns. ActivityNet-style datasets \cite{caba2015activitynet} extend this setting toward temporal localization and event-level understanding in longer, more complex videos. These benchmarks are fundamental for learning spatiotemporal representations, but their semantic target is usually explicit: the model must identify what action or event is visible in the clip. In contrast, \bmname{} evaluates a different layer of meaning: what message the clip communicates beyond the observable action or event.

\paragraph{Captioning and video--text retrieval.}
Video--language datasets such as MSVD~\cite{chen2011collecting}, MSR-VTT~\cite{xu2016msrvtt}, VATEX~\cite{wang2019vatex}, ActivityNet Captions~\cite{krishna2017densecaptions}, and LSMDC~\cite{rohrbach2016movie} evaluate the alignment between video content and natural language descriptions. These resources support captioning and retrieval tasks in which the text typically describes visible entities, actions, scenes, or events. Similarly, video--text retrieval methods such as CLIP4Clip~\cite{luo2022clipclip}, Frozen-in-Time~\cite{bain2021frozen}, VideoCLIP~\cite{xu2021videoclip}, X-CLIP~\cite{ni2022xclip}, ALPRO~\cite{li2022alpro}, InternVideo~\cite{wang2024internvideo2} and VideoPrism~\cite{zhaovideoprism} learn representations that match videos to caption-like text. Such models can perform well when retrieval depends on salient objects or actions, but they may fail when the query refers to an abstract message that is only implied. \bmname{} therefore evaluates retrieval under a more pragmatic criterion: matching clips and text by a holistic communicative meaning rather than by literal visual correspondence.

\paragraph{Video QA and temporal reasoning.}
Video question-answering benchmarks broaden the evaluation target from recognition to reasoning. TGIF-QA~\cite{jang2017tgifqa} tests spatiotemporal understanding over short clips, ActivityNet-QA~\cite{yu2019activitynetqa} evaluates question answering over web videos, and NExT-QA~\cite{xiao2021nextqa} emphasizes causal and temporal explanations. IntentQA~\cite{li2023intentqa} moves closer to high-level reasoning by asking about intentions in video situations. However, these datasets typically focus on events, causes, temporal relations, or the intentions of agents within the depicted scene. \bmname{} targets a different notion of intent: the communicative intent of the clip itself. The question is not only what an actor is doing or why an event happened, but what broader message the video, as a communicative artifact, encourages the viewer to infer.

\paragraph{Benchmarks for video MLLMs.}
Recent multimodal large language model benchmarks evaluate broader video understanding capabilities. MVBench~\cite{li2024mvbench} tests diverse video understanding tasks designed to reduce single-frame bias, while Video-MME~\cite{fu2025videomme} evaluates models across video durations, domains, and modalities such as subtitles and audio. VideoVista~\cite{li2024videovista}, TempCompass~\cite{liu2024tempcompass}, and the Perception Test~\cite{patraucean2023perceptiontest} probe different forms of visual, temporal, and multimodal reasoning. Long-form benchmarks such as EgoSchema~\cite{mangalam2023egoschema}, CinePile~\cite{rawal2024cinepile}, LVBench~\cite{wang2024lvbench}, and MLVU~\cite{zhou2025mlvu} further stress extended context, long-horizon dependencies, and multi-task comprehension. These benchmarks provide broad evaluations of perception, temporal reasoning, and long-context video understanding. \bmname{} is complementary: rather than expanding duration or task coverage, it isolates a specific challenge in short videos, namely whether models can infer communicative messages from often implicit cues.

\paragraph{High-level and domain-specific reasoning.}
Several recent benchmarks show the value of evaluating high-level social or domain-specific video semantics. SIV-Bench~\cite{kong2025sivbench} focuses on social interaction understanding, while AdsQA~\cite{long2025adsqa} studies advertisement video understanding, where persuasive intent and cognitive-level reasoning are central. These benchmarks move beyond low-level perception, but they remain tied to specific domains: social interaction in one case and advertising in the other. \bmname{} instead studies implicit message understanding in everyday short, shareable clips across multiple practical topics, including career and finance, education, health and well-being, culture, safety, sustainability, and lifestyle. This makes the benchmark closer to general short-form video interpretation, where messages are often implied through subtle multimodal or cultural cues.

\paragraph{How \bmname{} differs.}
Current video benchmarks often test recognition, caption-like alignment, temporal reasoning, factual question answering, or domain-specific inference. \bmname{} targets a missing semantic layer: the implicit message a short video is trying to communicate. It differs from prior work in four main ways. First, it focuses on short, internet-native clips with dense communicative content. Second, it is message-centric: the target is not an object, action, event, caption, or actor intention, but the high-level communicative proposition implied by the clip. Third, it requires multimodal grounding, since the intended message may depend on visual evidence, speech, text overlays, audio, temporal structure, and contextual cues. Fourth, it supports both multiple-choice message understanding and bidirectional message--clip retrieval, enabling evaluation of whether models can represent latent meaning at scale. 
As a result, \bmname{} evaluates whether models can recover what videos mean, not only what they depict.

\section{Data Collection}
\label{sec:data collection}
We focus on short, internet-native clips, using YouTube as a scalable and reproducible source of public web videos. Although our motivation extends to short-form platforms more broadly, the current version of \bmname{} should be interpreted as a YouTube-derived diagnostic benchmark for implicit message understanding.

\begin{figure}[t]
    \vspace{-7mm}
    \centering
    \includegraphics[width=\linewidth]{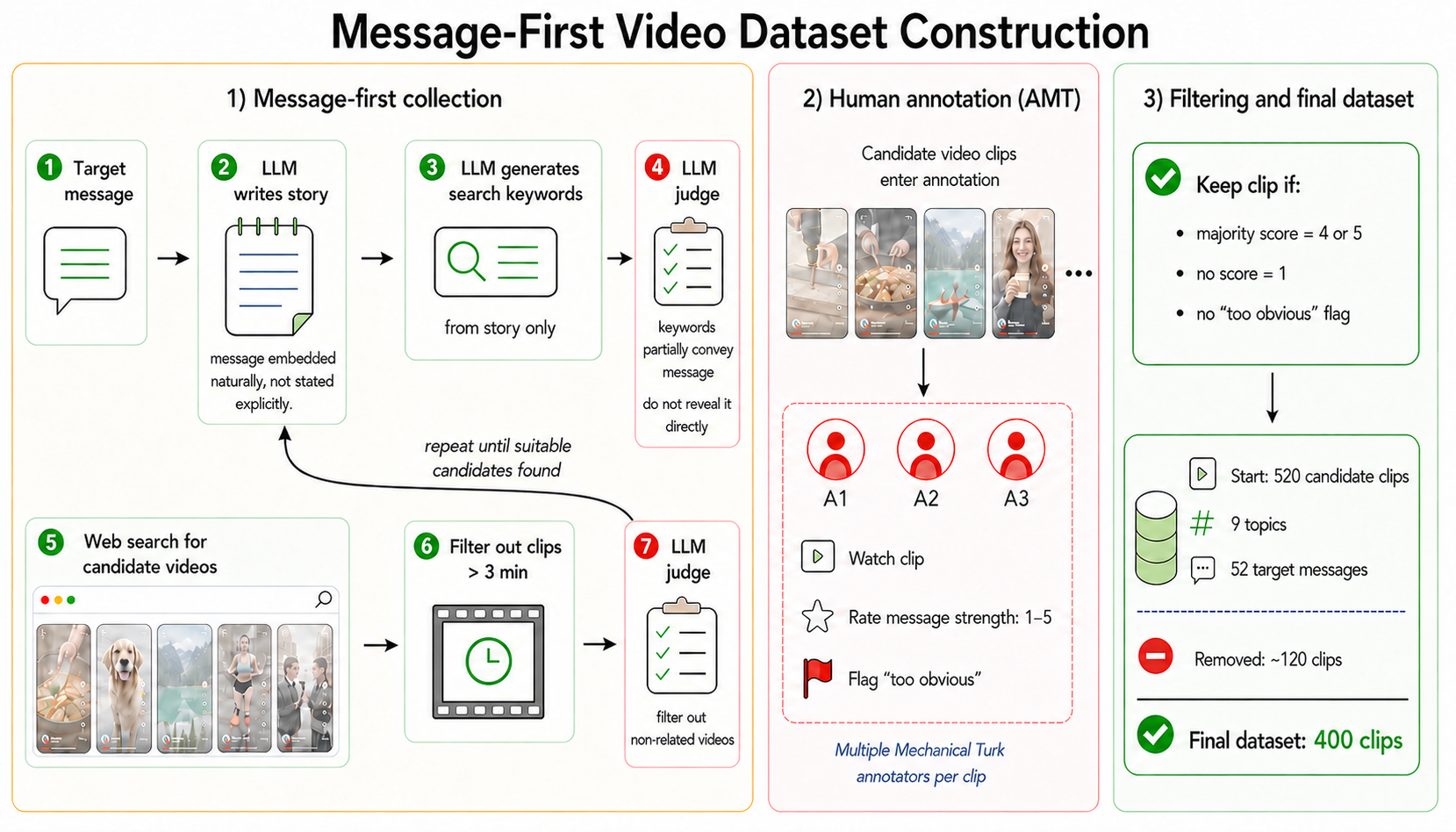}
    \captionsetup{font=small}
    \caption{Overview of the message-first video dataset construction pipeline. (1) Starting with a target message, (2) we prompt an LLM to generate a visual storyline in which the message is embedded implicitly, (3) An LLM generates indirect search keywords, (4) which are checked by one LLM judge and (5) used to retrieve candidate videos; (6) after duration filtering, (7) a second LLM judge removes unrelated clips, with the process repeated until a suitable candidate is found. The clips are then annotated by three AMT workers and conservatively filtered, yielding 400 clips from 520 candidates.
    }
    \label{fig:data-collection}
    \vspace{-2mm}
\end{figure}

Constructing a benchmark for implicit message understanding is challenging because standard web search is optimized for explicit objects, events, keywords, and metadata. Directly searching for a target message often retrieves clips in which the message is {\it explicitly} stated verbatim in the title, speech, caption, or text overlay. Such examples are useful for ordinary video--text matching but are less suitable for evaluating whether models can infer a message from indirect multimodal evidence.

To address this issue, we adopt the message-first collection pipeline shown in Fig.~\ref{fig:data-collection}. The process starts with a target message. An LLM is then prompted to generate short, story-like scenarios that naturally incorporate the target message into the narrative. From these stories, a second LLM generates indirect search keywords based on the story, and a separate LLM judge verifies that the keywords remain semantically related to the original message while avoiding direct lexical disclosure. We use the resulting keywords to retrieve candidate public YouTube clips and discard clips longer than three minutes. This process is repeated until a sufficient number of candidate clips is obtained for each target message.

Candidate clips are then validated through human annotation on Amazon Mechanical Turk. Each clip is assigned to three independent annotators. Annotators watch the clip and rate, on a five-point scale, how strongly it conveys the target message. They are also asked to flag clips in which the message is conveyed too explicitly, for example when a speaker states the target message directly or when it appears as prominent text on screen. The annotation interface is shown in Fig.~\ref{fig:amt-screenshot} in the Appendix.

\begin{figure}[b]
    \centering
    \includegraphics[width=0.95\linewidth]{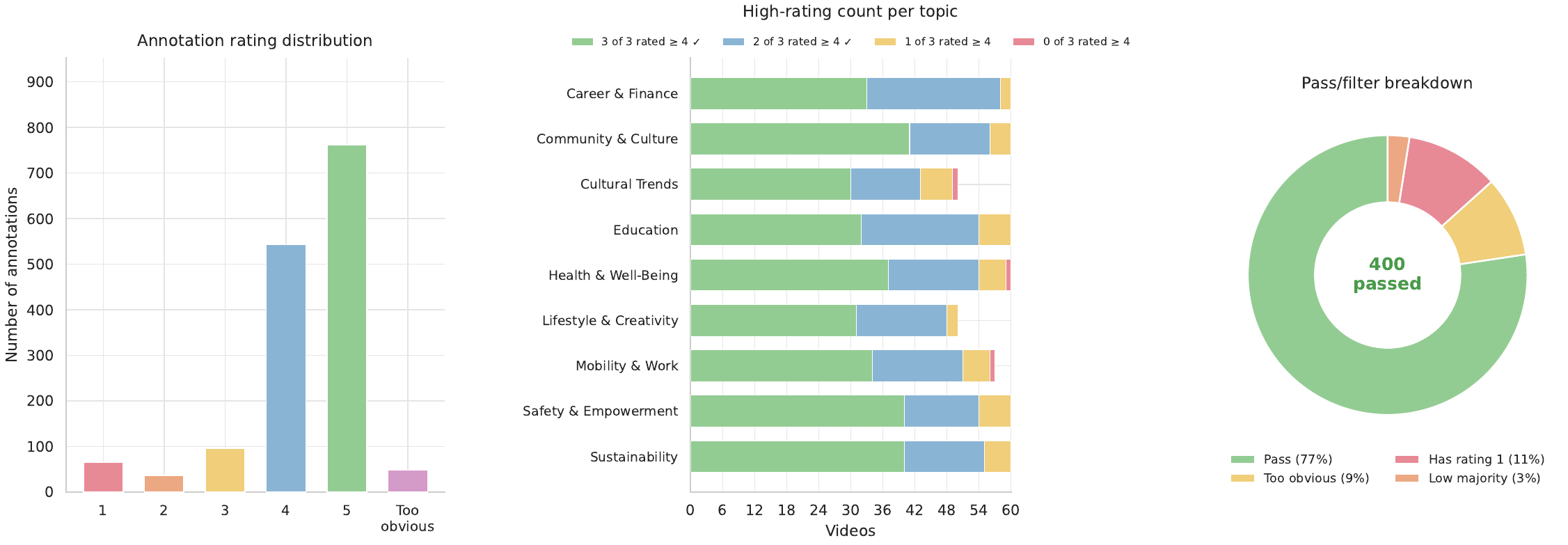}
    \caption{
    Annotation score distribution and filtering outcomes. Clips are retained only when they satisfy the majority high-relevance rule and are not flagged as either irrelevant or too explicit.
    }
    \label{fig:annotation}
\end{figure}

We initially collected 520 candidate clips covering 9 topics and 52 target messages. Fig.~\ref{fig:Schematic Example and Clip Duration} indicates schematic examples and clip duration statistics, while Fig.~\ref{fig:VidMsg Topic Distribution} shows the topic distribution (see also Sec. \ref{sec:data statistics}). To ensure that retained clips are both relevant and non-trivial, we apply a conservative filtering rule. A clip is retained only if a majority of annotators (among three) assign a high relevance score, i.e., 4 or 5, no annotator assigns a score of 1, and no annotator marks the clip as ``message too obvious.'' This filtering removes approximately 120 candidates and yields a final benchmark of 400 clips. This corresponds to a 77\% success rate in our automatic data collection strategy. The distribution of annotation scores and filtering outcomes is shown in Fig.~\ref{fig:annotation}.

The resulting 77\% retention rate suggests that the message-first retrieval pipeline is effective at surfacing clips that are semantically related to the target messages. At the same time, the human filtering stage removes clips that are either weakly related or overly explicit, ensuring that \bmname{} focuses on examples where the intended message must be inferred from the clip rather than copied from direct textual or spoken evidence. Annotation instructions are provided in the Appendix Fig.~\ref{fig:amt-screenshot}.

\section{Data Statistics}
\label{sec:data statistics}
Fig.~\ref{fig:Schematic Example and Clip Duration} shows a few examples from \bmname{} together with the clip-duration distribution. Most clips are short, with the majority lasting under one minute, while the longest retained clips are below three minutes. This reflects our focus on short, internet-native videos that are likely to contain dense communicative content.

Fig.~\ref{fig:VidMsg Topic Distribution} summarizes the distribution of clips across topics and target messages. \bmname{} covers nine recurring communicative domains: career and finance, community and culture, cultural trends, education, health and well-being, lifestyle and creativity, mobility and work, safety and empowerment, and sustainability. These topics were selected because clips in such domains often convey abstract advice, values, emotions, social meanings, or implicit viewpoints rather than merely documenting visible events.

The dataset is relatively balanced across topics. As shown in Fig.~\ref{fig:VidMsg Topic Distribution}(a), the number of clips per topic ranges from 35 to 49, with the largest topics being community and culture, safety and empowerment, and sustainability. Fig.~\ref{fig:VidMsg Topic Distribution}(b) shows the distribution of clips per target message. Across 52 target messages, each message is represented by 5 to 9 clips, with most messages having 8 or 9 clips. This structure supports both broad topic coverage and fine-grained message-level evaluation.

Importantly, \bmname{} is designed to reduce reliance on coarse topic recognition. Within each topic, the benchmark includes semantically related but distinct messages, such as ``food as self-care'' versus ``lifestyle medicine beyond symptom relief,'' or ``work-life integration'' versus ``performance over work hours.'' As a result, models must distinguish fine-grained communicative meanings rather than simply identify the general domain of the clip.

\paragraph{Annotation Agreement.}
We evaluate inter-rater reliability using average pairwise linear-weighted $\kappa$ among the three annotators assigned to each clip. Since annotators could also flag a clip as ``too obvious,'' we encode this option as an additional positive category (6) when computing agreement. On a collapsed relevance scale, where ratings 4--6 are treated as positive, rating 3 as neutral, and ratings 1--2 as negative, agreement is substantial across all annotated clips ($\kappa = 0.681$).

Agreement is higher among the 400 retained clips ($\kappa = 0.795$), which is expected because the filtering rule keeps only clips with majority high-relevance judgments and removes clips with strong disagreement or explicit-message flags. These results indicate that annotators generally agree on whether a clip conveys the target message. At the same time, some variation in rating intensity is expected, since implicit message understanding is inherently nuanced and may depend on how strongly each annotator perceives the clip's communicative intent.
\section{Evaluation}
\label{sec:evaluation}
We evaluate \bmname{} primarily in a bidirectional message--clip retrieval setting, where the goal is to match target messages and clips according to the implicit meaning conveyed by the video. In text-to-video retrieval, each of the 52 target messages is used as a query, and the model must retrieve the relevant clips from the full pool of 400 clips. In video-to-text retrieval, each clip is used as a query, and the model must retrieve the corresponding target message from the set of 52 messages. This setting reflects the main motivation of \bmname{}: enabling scalable search, clustering, recommendation, moderation, and analysis of videos by what they communicate rather than only by what they depict.

We report ranking performance using mean Average Precision (mAP). Many off-the-shelf video foundation models process visual frames separately from text and do not natively support joint audio or transcribed speech inputs; for these models, evaluation follows their standard input protocol.

\subsection{\ours{} Baseline Method}

Collecting large-scale training data for message--clip alignment is costly, since implicit messages require human interpretation and careful filtering. To provide a strong retrieval baseline, we extend the text-only optimization strategy introduced by \citet{vidvec2026} to the message-retrieval setting. The key idea is to adapt a video MLLM toward message-oriented representations without requiring additional video training data.

Specifically, we generate synthetic clip-storyline--message pairs that capture the relationship between an implicit communicative message and the visual storyline that conveys it. To encourage diversity, we sample personas from PersonaHub~\cite{personahub} and prompt an LLM to generate, for each persona, both a short target message and a dense visual storyline from which the message is inferable. Following \citet{vidvec2026}, we use fewer than 70K synthetic clip-storyline--message pairs to optimize an MLLM. This lightweight text-only optimization encourages the model to attend to contextual visual cues that may imply a broader communicative message, rather than treating the task as ordinary caption matching.

\subsection{Retrieval Results}

\begin{table}[t]
\centering
\small
\caption{\bmname{} retrieval performance comparison using Recall@10, Recall@1, and mAP on text-to-video and video-to-text tasks, with and without transcripts.}
\begin{tabular}{lcccccccc}
\toprule
\multirow{3}{*}{Method} 
& \multicolumn{4}{c}{Text-to-Video} 
& \multicolumn{4}{c}{Video-to-Text} \\
\cmidrule(lr){2-5} \cmidrule(lr){6-9}
& \multicolumn{2}{c}{w/o trans} 
& \multicolumn{2}{c}{w trans} 
& \multicolumn{2}{c}{w/o trans} 
& \multicolumn{2}{c}{w trans} \\
\cmidrule(lr){2-3} \cmidrule(lr){4-5} \cmidrule(lr){6-7} \cmidrule(lr){8-9}
& R@10 & mAP & R@10 & mAP & R@1 & mAP & R@1 & mAP \\
\midrule
Clip4Clip~\cite{luo2022clipclip}            & 21.8 & 19.3 & -- & --    & 24.1  & 37.3 & -- & --    \\
ImageBind~\cite{girdhar2023imagebind}        & 24.6 & 23.6 & -- & --    & 20.0  & 34.8 & -- & --    \\
PE-Core~\cite{bolya2025PerceptionEncoder}    & 10.3 & 10.0 & -- & --    & 8.1   & 16.8 & -- & --    \\
InternVideo2-1B~\cite{wang2024internvideo2}  & 19.5 & 17.4 & -- & --    & 17.7  & 29.4 & -- & --    \\
InternVideo2-6B~\cite{wang2024internvideo2}  & 25.5 & 22.6 & -- & --    & 23.0  & 36.1 & -- & --    \\
VideoPrism~\cite{zhaovideoprism}             & 28.4 & 24.8 & -- & --    & 29.9  & 42.8 & -- & --    \\
VLM2Vec-2.0~\cite{meng2025vlm2vecv2}         & 29.0 & 27.6 & 32.3 & 31.0 & 27.9 & 43.6 & 34.4 & 49.2 \\
Qwen3-VL-Emb~\cite{li2026qwen3vlemb}         & 39.7 & 36.1 & 42.7 & 42.8 & 45.8 & 50.6 & \textbf{50.6} & 63.1 \\
VidVec~\cite{vidvec2026}                     & 28.4 & 25.3 & 32.1 & 28.6 & 36.0  & 51.0 & 39.8 & 54.5 \\
\midrule
\ours                                        & \textbf{47.5} & \textbf{45.6} & \textbf{48.9} & \textbf{46.7} & \textbf{48.9} & \textbf{63.2} & 49.1 & \textbf{65.0} \\
\bottomrule
\end{tabular}
\label{tab:t2v_v2t_map_results}
\end{table}

Table~\ref{tab:t2v_v2t_map_results} compares \ours{} with video retrieval and multimodal embedding baselines on text-to-video and video-to-text retrieval, with and without transcripts. We report Recall@10 and mAP for text-to-video, and Recall@1 and mAP for video-to-text. Overall, \bmname{} is challenging for standard retrieval models: methods optimized for visible entities, actions, or caption-like alignment, such as Clip4Clip~\cite{luo2022clipclip}, PE-Core~\cite{bolya2025PerceptionEncoder}, and InternVideo2~\cite{wang2024internvideo2}, perform relatively poorly. VideoPrism~\cite{zhaovideoprism} improves over weaker baselines but still lags behind message-oriented models, suggesting that strong visual representations alone are insufficient for abstract message inference.

Among off-the-shelf baselines, Qwen3-VL-Emb~\cite{li2026qwen3vlemb} performs best for text-to-video retrieval, reaching \(39.7\) Recall@10 and \(36.1\) mAP without transcripts, and \(42.7\) Recall@10 and \(42.8\) mAP with transcripts. For video-to-text, Qwen3-VL-Emb achieves the highest Recall@1, while VidVec~\cite{vidvec2026} obtains the strongest mAP without transcripts. Transcript input improves supported models, indicating that speech cues are useful for message-level retrieval, but the remaining gap to \ours{} shows that general MLLM embeddings, even when augmented with transcripts, do not fully capture implicit communicative meaning.

\ours{} achieves the strongest overall performance. Without transcripts, it outperforms the best text-to-video baseline, Qwen3-VL-Emb~\cite{li2026qwen3vlemb}, by \(7.8\) Recall@10 and \(9.5\) mAP points, and improves video-to-text retrieval by \(3.1\) Recall@1 and \(12.2\) mAP points over the strongest baselines. With transcripts, \ours{} achieves the best text-to-video results and the highest video-to-text mAP (\(65.0\)). Qwen3-VL-Emb slightly leads on transcript-based video-to-text Recall@1, while \ours{} achieves better overall ranking as reflected by mAP.

The results also highlight why \bmname{} differs from conventional video retrieval benchmarks. In \bmname{}, the evaluation target shifts from matching what is shown to matching what is communicated. Textual cues help when they contain information related to the intended message. At the same time, the modest transcript gains for \ours{} suggest that much of the relevant signal is already captured through visual and contextual message-oriented alignment. Overall, the strongest performance comes from adapting the embedding space toward implicit communicative meaning.

\subsection{Multiple-Choice Question Answering}
\label{sec:mcq}

\begin{table}[t]
\vspace{-5mm}

\small
\centering
\setlength{\tabcolsep}{4pt}
\renewcommand{\arraystretch}{1.1}
\caption{VidMsg QA accuracy (\%) by topic. Topics include Career, Community, Cultural, Education, Health, Lifestyle, Mobility, Safety, and Sustainability. Models are sorted by parameter count within each model group. The best overall result and the best open-source result in each column are bolded.}
\label{tab:qa_results}

\begin{tabular}{l c c c c c c c c c|c}
\toprule
\textbf{Model}
& \textbf{Car.}
& \textbf{Comm.}
& \textbf{Cult.}
& \textbf{Edu.}
& \textbf{Hlth.}
& \textbf{Life.}
& \textbf{Mob.}
& \textbf{Safe.}
& \textbf{Sust.}
& \textbf{Overall} \\
\midrule
\rowcolor{gray!12}
\multicolumn{11}{c}{\textbf{Commercial Large Multimodal Models}} \\
\midrule
GPT-5.4-Mini-2026-03-17
& 46.5 & 73.5 & 60.0 & 65.2 & 73.3 & 71.8 & 65.9 & 76.6 & \textbf{97.9} & 70.6 \\
GPT-5.4-2026-03-05
& 55.8 & 71.4 & 54.3 & 67.4 & 75.6 & 69.2 & 65.9 & 72.3 & 95.7 & 70.4 \\
Gemini-3-Flash
& \textbf{67.4} & 75.5 & 65.7 & 73.9 & 73.3 & 74.4 & 81.8 & 76.6 & 95.7 & \textbf{76.5} \\
Gemini-3.1-Pro
& 65.1 & 67.3 & 65.7 & \textbf{80.4} & 75.6 & 71.8 & \textbf{84.1} & \textbf{78.7} & 95.7 & \textbf{76.5} \\

\midrule
\rowcolor{gray!12}
\multicolumn{11}{c}{\textbf{Open-source Vision-Language Models}} \\
\midrule
Qwen3-VL-2B~\cite{bai2025qwen3vl}
& 41.9 & 67.3 & 65.7 & 54.4 & 57.8 & 61.5 & 65.9 & 66.0 & 95.7 & 64.3 \\
VideoLLaMA3-7B~\cite{damonlpsg2025videollama3}
& 27.9 & 40.8 & 31.4 & 39.1 & 33.3 & 46.1 & 31.8 & 48.9 & 61.7 & 40.5 \\
Qwen2.5-VL-7B~\cite{bai2025qwen25}
& \textbf{55.8} & 77.5 & \textbf{68.6} & \textbf{63.0} & 68.9 & 69.2 & \textbf{79.5} & \textbf{70.2} & 91.5 & \textbf{71.9} \\
MiniCPM-V-4.5-8B~\cite{yu2025minicpmv45cookingefficient}
& 20.9 & 22.4 & 28.6 & 17.4 & 22.2 & 25.6 & 20.4 & 23.4 & 21.3 & 22.3 \\
Molmo2-8B~\cite{molmo2openweightsdata}
& 44.2 & 55.1 & 40.0 & 56.5 & 48.9 & 61.5 & 45.5 & \textbf{72.3} & 89.4 & 57.7 \\
NVILA-8B~\cite{liu2025nvila}
& 41.9 & 55.1 & 57.1 & 45.6 & 48.9 & 46.1 & 40.9 & 66.0 & 78.7 & 53.7 \\
Qwen3-VL-8B~\cite{bai2025qwen3vl}
& 53.5 & 61.2 & 60.0 & 47.8 & 64.4 & 69.2 & 61.4 & 68.1 & 95.7 & 64.8 \\
LLaVA-1.5-13B~\cite{llava15}
& 46.5 & 57.1 & 60.0 & 47.8 & 46.7 & 59.0 & 50.0 & 48.9 & 80.8 & 55.2 \\
Qwen2.5-VL-32B~\cite{bai2025qwen25}
& 46.5 & \textbf{79.6} & 48.6 & 50.0 & \textbf{77.8} & 66.7 & 65.9 & 70.2 & 95.7 & 67.6 \\
Qwen3-VL-32B~\cite{bai2025qwen3vl}
& 53.5 & 75.5 & 57.1 & 58.7 & 66.7 & \textbf{76.9} & 65.9 & 63.8 & \textbf{97.9} & 68.9 \\

\bottomrule
\end{tabular}
\end{table}

Although retrieval is the primary evaluation setting in \bmname{}, we also include a multiple-choice question-answering (MCQ) setting as a diagnostic probe of implicit message understanding. Unlike retrieval, which evaluates whether models can encode messages and clips into a shared representation, the MCQ setting evaluates MLLMs in their native instruction-following mode. For each clip, the model is asked to select the main message conveyed by the video from five candidate messages. All candidates are sampled from the same topic using a fixed random seed, with exactly one correct answer. This same-topic design makes the task more challenging than coarse topic classification, since the distractors are semantically related to the correct message.

Table~\ref{tab:qa_results} reports MCQ accuracy across topics and overall. Among open-source models, Qwen2.5-VL-7B achieves the best overall performance, reaching \(71.9\%\). It outperforms larger open-source variants such as Qwen2.5-VL-32B (\(67.6\%\)) and Qwen3-VL-32B (\(68.9\%\)), suggesting that model scale alone does not guarantee better message-level reasoning. Within the Qwen3-VL family, however, performance increases with scale, from \(64.3\%\) at 2B to \(68.9\%\) at 32B.

Commercial models obtain the strongest overall results. Gemini-3-Flash and Gemini-3.1-Pro both reach \(76.5\%\), outperforming all open-source models. This advantage may partly reflect Gemini's native support for video processing. GPT-5.4-mini and GPT-5.4 achieve similar performance, \(70.6\%\) and \(70.4\%\), respectively, comparable to the strongest open-source models but below the Gemini models. These results indicate that current MLLMs show meaningful progress in identifying the intended message when given a small set of semantically related options, but the task remains far from saturated.

The per-topic results reveal substantial variation in difficulty. Sustainability is consistently the easiest topic, with several models exceeding \(95\%\) accuracy, suggesting that its messages may be more visually or contextually distinctive. In contrast, topics such as career and finance, cultural trends, and lifestyle and creativity are more challenging for many models, likely because they require finer pragmatic distinctions and stronger contextual interpretation. 

Several models perform substantially below the strongest systems. VideoLLaMA3-7B achieves only \(40.5\%\), and MiniCPM-V-4.5-8B obtains \(22.3\%\), close to the \(20\%\) random-choice baseline.
These results show that not all video-oriented or general vision-language models can reliably infer implicit communicative messages, even in a constrained five-choice setting.
Overall, the MCQ results complement the retrieval evaluation by showing that \bmname{} is not merely a representation-learning challenge: it also exposes gaps in direct message reasoning, especially when the correct answer must be selected among closely related alternatives.

\subsection{Confusion Analysis}
\label{sec:confusion}

\begin{figure}[t]
\vspace{-8mm}
    \centering

    \begin{subfigure}{0.53\linewidth}
        \centering
        \includegraphics[width=0.95\linewidth]{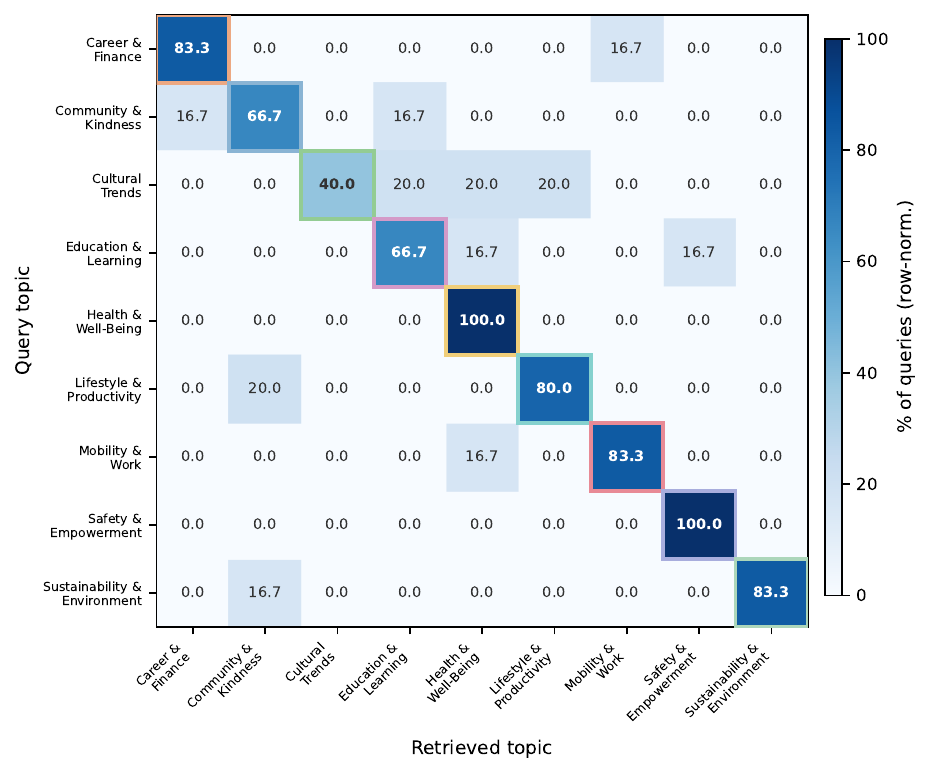}
        \caption{Topic-level retrieval confusion matrix for Qwen3-VL-Embedding. Rows indicate query topics and columns indicate retrieved topics; percentage of top-1 retrievals.}
        \label{fig:qwen3vl-topic-retrieval-confusion}
    \end{subfigure}
    \hfill
    \begin{subfigure}{0.45\linewidth}
        \centering
        \includegraphics[width=\linewidth]{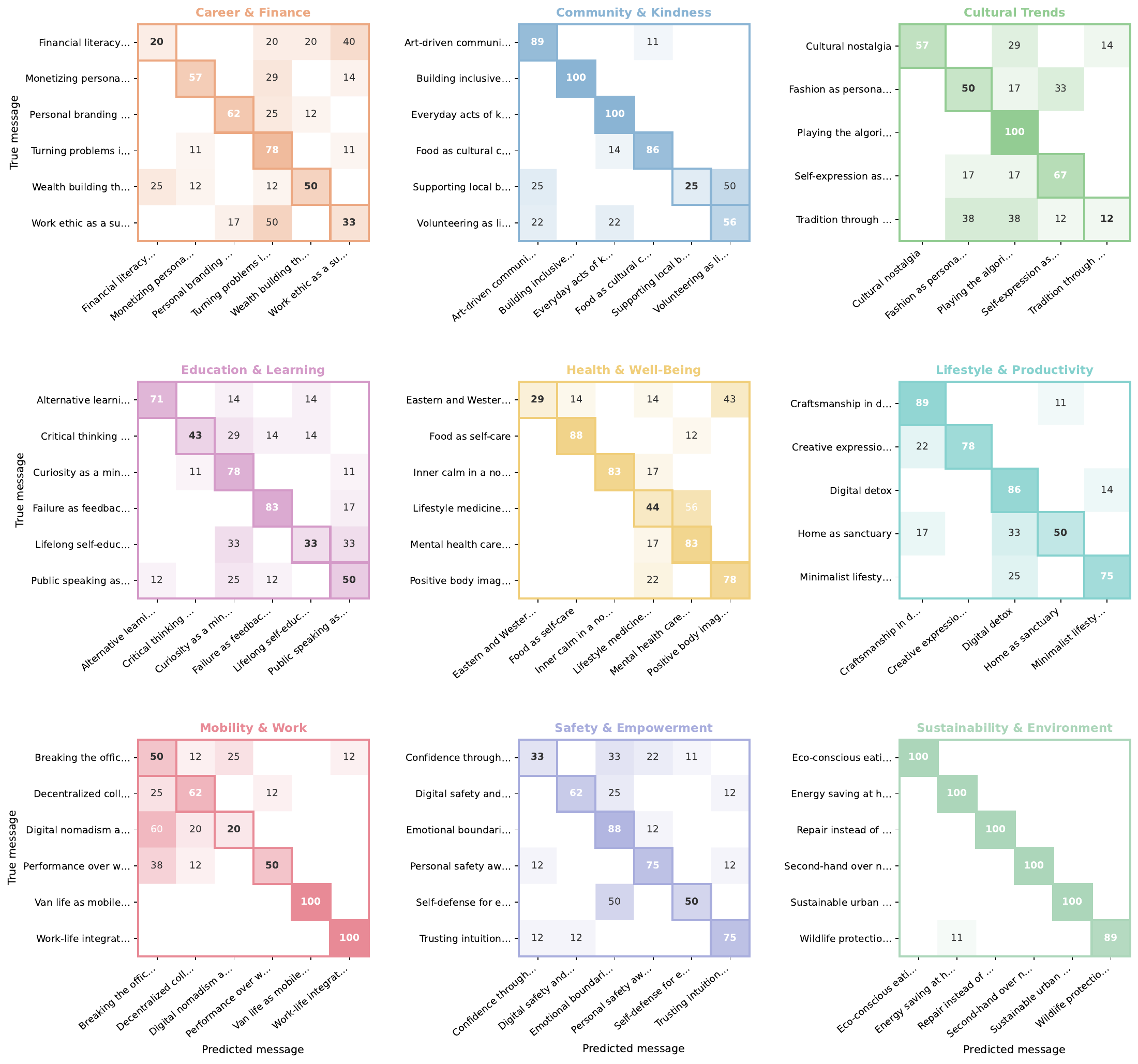}
        \caption{Within-topic message-level confusion matrices for Qwen3-VL-32B QA. Each block shows the distribution of predicted messages within the topic.}
        \label{fig:qwen3vl-intratopic-confusion}
    \end{subfigure}

    \caption{Qwen3-VL confusion analysis.}
    \label{fig:qwen3vl-confusion-analysis}
    \vspace{-4mm}
\end{figure}

To better understand where models fail, we analyze confusion patterns at both the topic and message levels. Fig.~\ref{fig:qwen3vl-topic-retrieval-confusion} shows the topic-level top-1 retrieval confusion matrix for Qwen3-VL-Embedding. Overall, the model often retrieves clips from the correct topic, indicating that it captures coarse semantic domains reasonably well. Several topics show strong diagonal accuracy, including Health \& Well-Being and Safety \& Empowerment, both reaching \(100\%\), as well as Career \& Finance, Mobility \& Work, and Sustainability \& Environment, each reaching \(83.3\%\). However, the model is less reliable for more ambiguous or overlapping domains. Cultural Trends is the most difficult topic, with only \(40.0\%\) of queries retrieving a clip from the correct topic, and its errors spread across Education, Health \& Well-Being, Lifestyle \& Productivity, and related areas. This suggests that topic-level retrieval is easier when the visual and contextual cues are distinctive, but harder when the topic depends on cultural framing or abstract social meaning.

Fig.~\ref{fig:qwen3vl-intratopic-confusion} provides a finer-grained view by showing within-topic message-level confusion matrices for Qwen3-VL-32B in the MCQ setting. These results show that even when the topic is fixed, models often confuse semantically close messages. For example, within Career \& Finance, messages such as ``monetizing personal passions'', ``personal branding as career leverage'' and ``wealth building through consistency'' are frequently confused, reflecting their shared association with self-improvement and financial aspiration. Similarly, in Education \& Learning, the model confuses messages related to curiosity, critical thinking, lifelong learning, and alternative learning models. In Safety \& Empowerment, errors occur between messages such as ``digital safety and self-protection'', ``self-defense for empowerment'' and ``confidence through solo travel'', which all center on agency and protection.

These patterns indicate that \bmname{} is not solved by recognizing the broad topic of a clip. Models can often identify the general domain, but still fail to infer the specific communicative message intended by the video. This supports the design of \bmname{} as a message-level benchmark: the central challenge is not only to map clips to coarse categories, but to distinguish implicit meanings among related alternatives.

\section{Summary}
In this paper, we introduced a new benchmark for message in videos, focusing on cases where the relevant semantic content is implicit rather than directly observable or explicitly stated. We presented a message-first data collection pipeline, validated candidate clips through human annotation, and evaluated recent multimodal models under both retrieval and question-answering settings. The Appendix provides additional details on the benchmark construction, our \ours baseline, limitations, societal impact, and implementation. Overall, our results show that message retrieval remains challenging for existing models, highlighting the need for deeper inferential video understanding. We further introduced a strong baseline method, demonstrating that task-specific text-only optimization can substantially improve message--clip alignment without requiring direct video-based training.

\bibliography{neurips_2026}
\bibliographystyle{plainnat}

\appendix
\clearpage
\renewcommand{\thefigure}{A\arabic{figure}} 
\setcounter{figure}{0} 

\newtcblisting{promptbox}{
  enhanced,
  breakable,
  listing only,
  colback=white,
  colframe=black!60,
  boxrule=0.4pt,
  arc=2pt,
  left=5pt,
  right=5pt,
  top=5pt,
  bottom=5pt,
  drop shadow=black!25,
  listing options={
    basicstyle=\ttfamily\scriptsize,
    breaklines=true,
    columns=fullflexible,
    keepspaces=true
  }
}
\section*{Appendix}
This Appendix includes details on the following topics:
\begin{enumerate}[itemsep=2pt, topsep=2pt, parsep=0pt, partopsep=0pt]
  \item[\textbf{A.}] VidMsg Topics, Messages and Examples
  \item[\textbf{B.}] Message-First Video Dataset Construction
  \item[\textbf{C.}] Details on \ours
  \item[\textbf{D.}] Implementation Details
  \item[\textbf{E.}] Limitations
  \item[\textbf{F.}] Societal Impact
  \item[\textbf{G.}] Licenses
\end{enumerate}

\section{\bmname{} Topics, Messages and Examples}
\begin{figure}
    \centering
    \includegraphics[width=\linewidth]{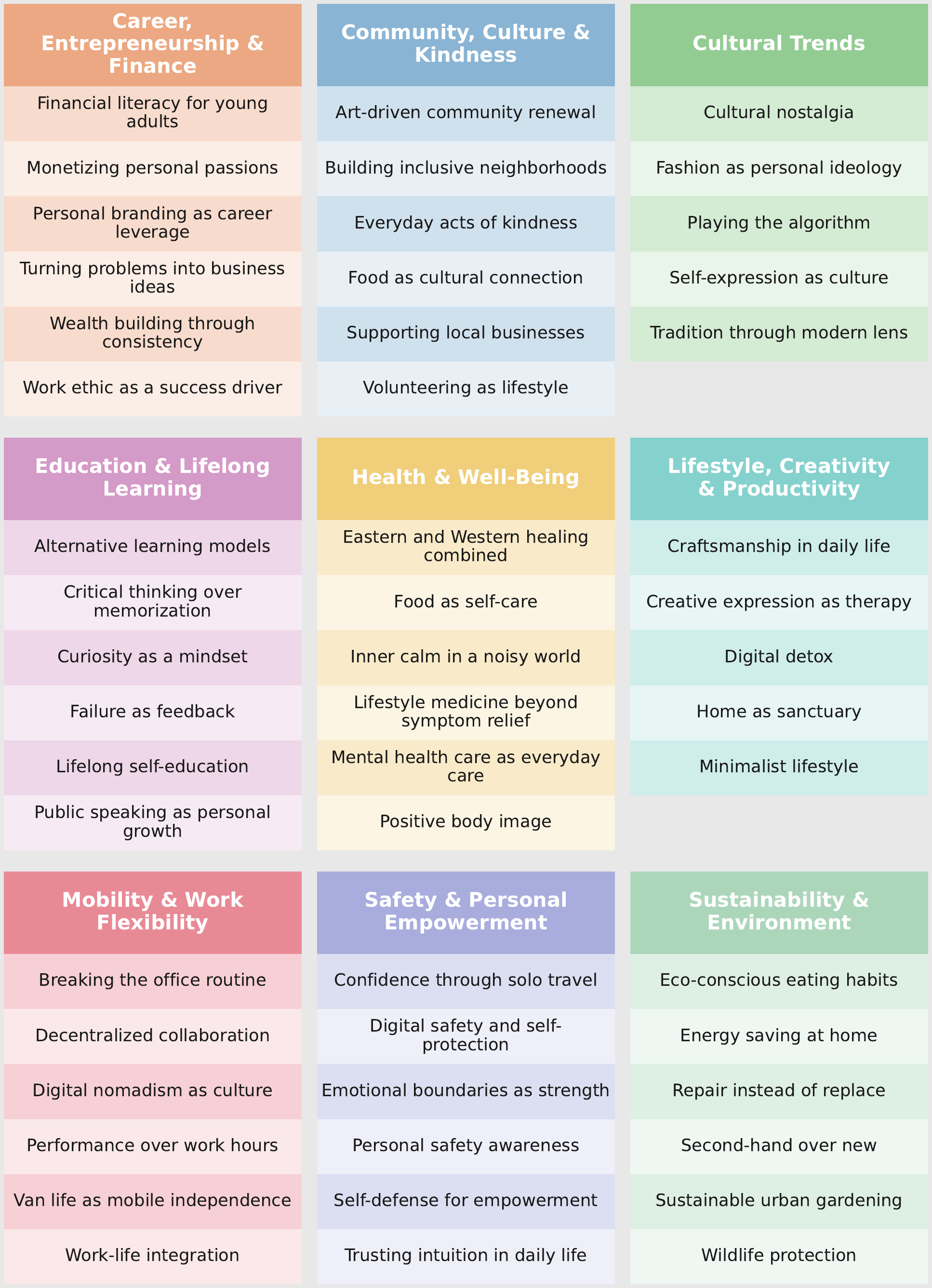}
    \caption{Overview of the nine topics and their corresponding messages in \bmname}
    \label{fig:alluvial}
\end{figure}

\begin{figure}
    \centering
    \includegraphics[width=\linewidth]{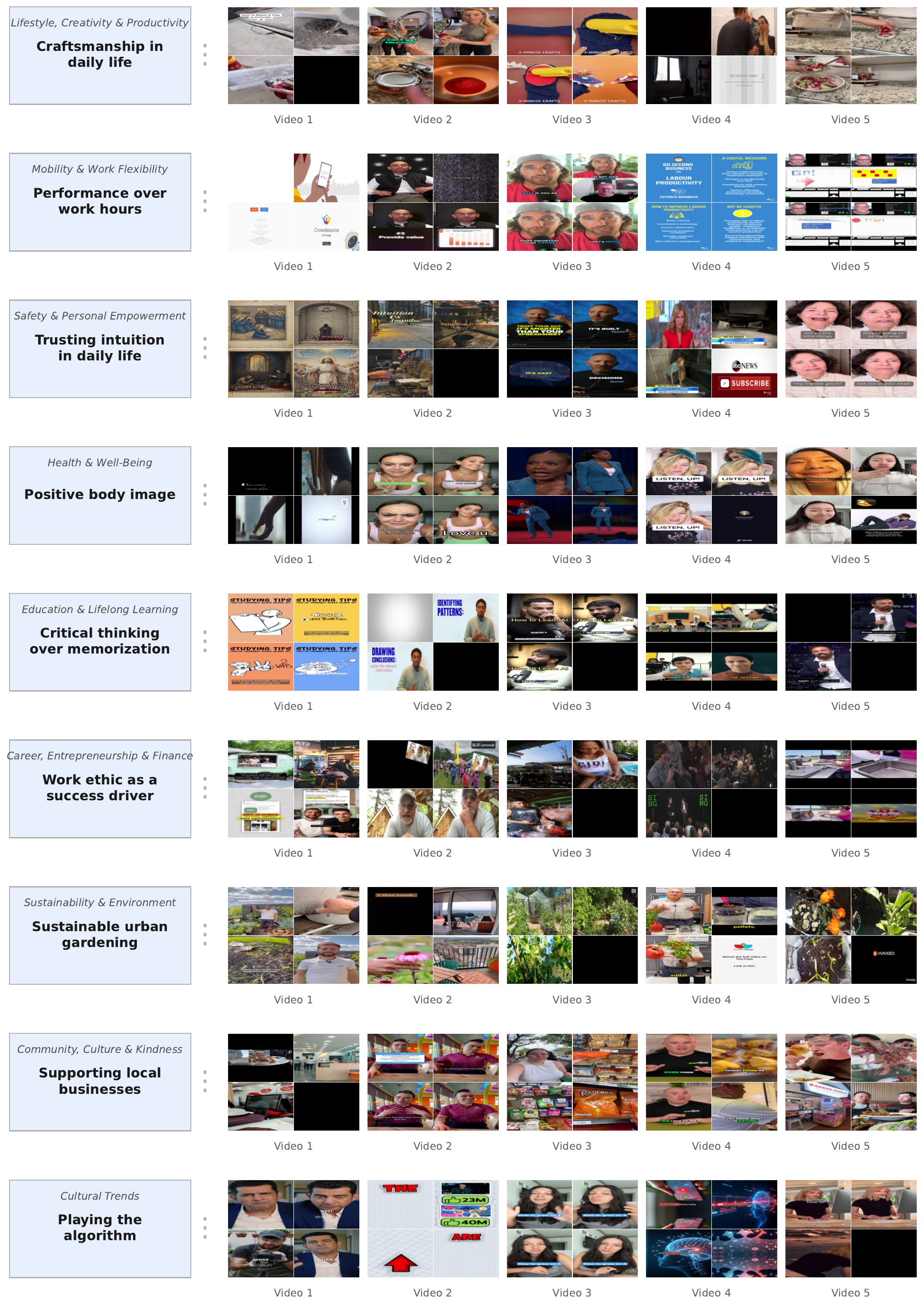}
    \caption{Example messages in \bmname, each paired with multiple video clips.}
    \label{fig:message-videos-examples}
\end{figure}

\begin{figure}
    \centering
    \includegraphics[width=\linewidth]{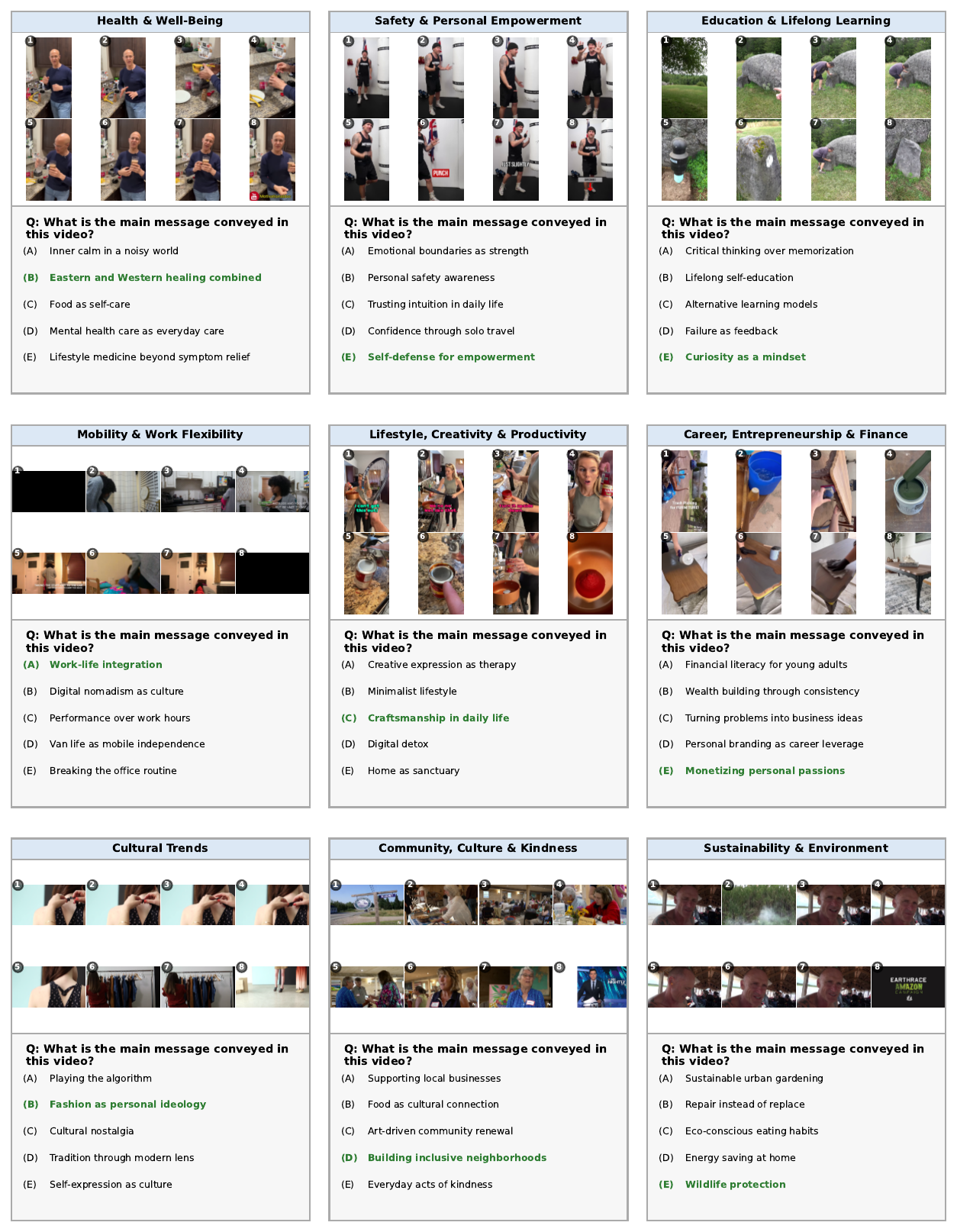}
    \caption{\bmname-QA: an example per topic.}
    \label{fig:mcq-examples}
\end{figure}

We provide a complete overview of \bmname{} messages and topics in Fig.~\ref{fig:alluvial}, which presents all 52 messages and their associated topics. We also provide qualitative examples in Fig.~\ref{fig:message-videos-examples}, showing one representative message per topic accompanied by thumbnails from five clips. Finally, Fig.~\ref{fig:mcq-examples} provides per-topic examples from \bmname{}-QA.

\section{Message-First Video Dataset Construction}
\label{sec:appendix-dataset-construction}
\subsection{Video Collection and Annotation Interfaces}

\begin{figure}[b]
    \centering
    \includegraphics[width=0.95\linewidth]{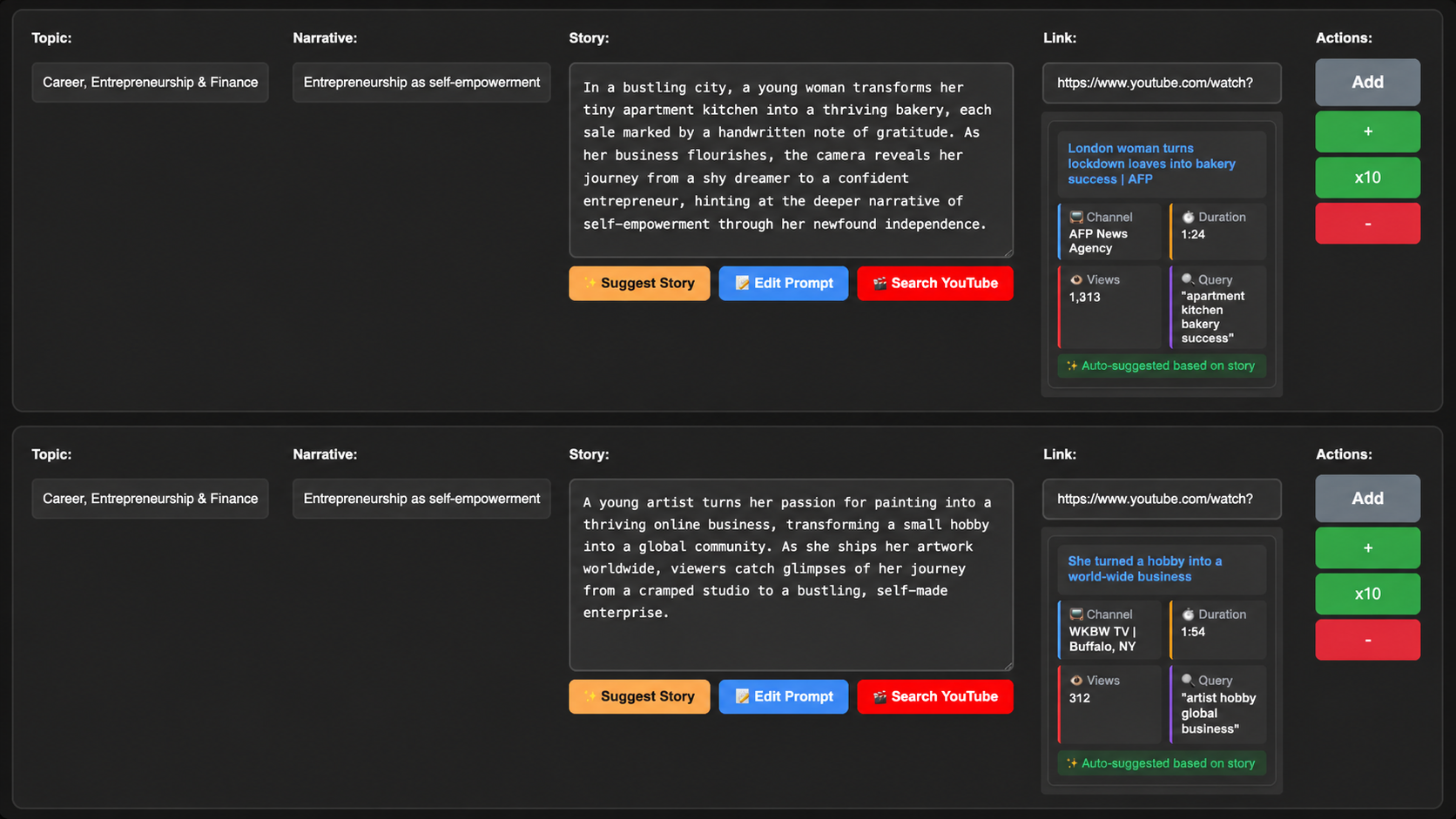}
    \captionsetup{font=small}
    \caption{Message-first video clip collection interface structured around repeated story-entry rows. Each row contains a generated story, generated search keywords, and a final candidate YouTube link.}
    \label{fig:vid-story-search-screenshot}
\end{figure}

Fig.~\ref{fig:vid-story-search-screenshot} shows the interface used during video clip collection. The interface is organized around message-first story entries: for each target message, the system presents a generated story, associated search keywords, and a final candidate YouTube link. This structure supports a consistent collection workflow in which candidate videos are selected according to their relevance to the intended message rather than through unconstrained video search.

\begin{figure}
    \centering
    \includegraphics[width=0.95\linewidth]{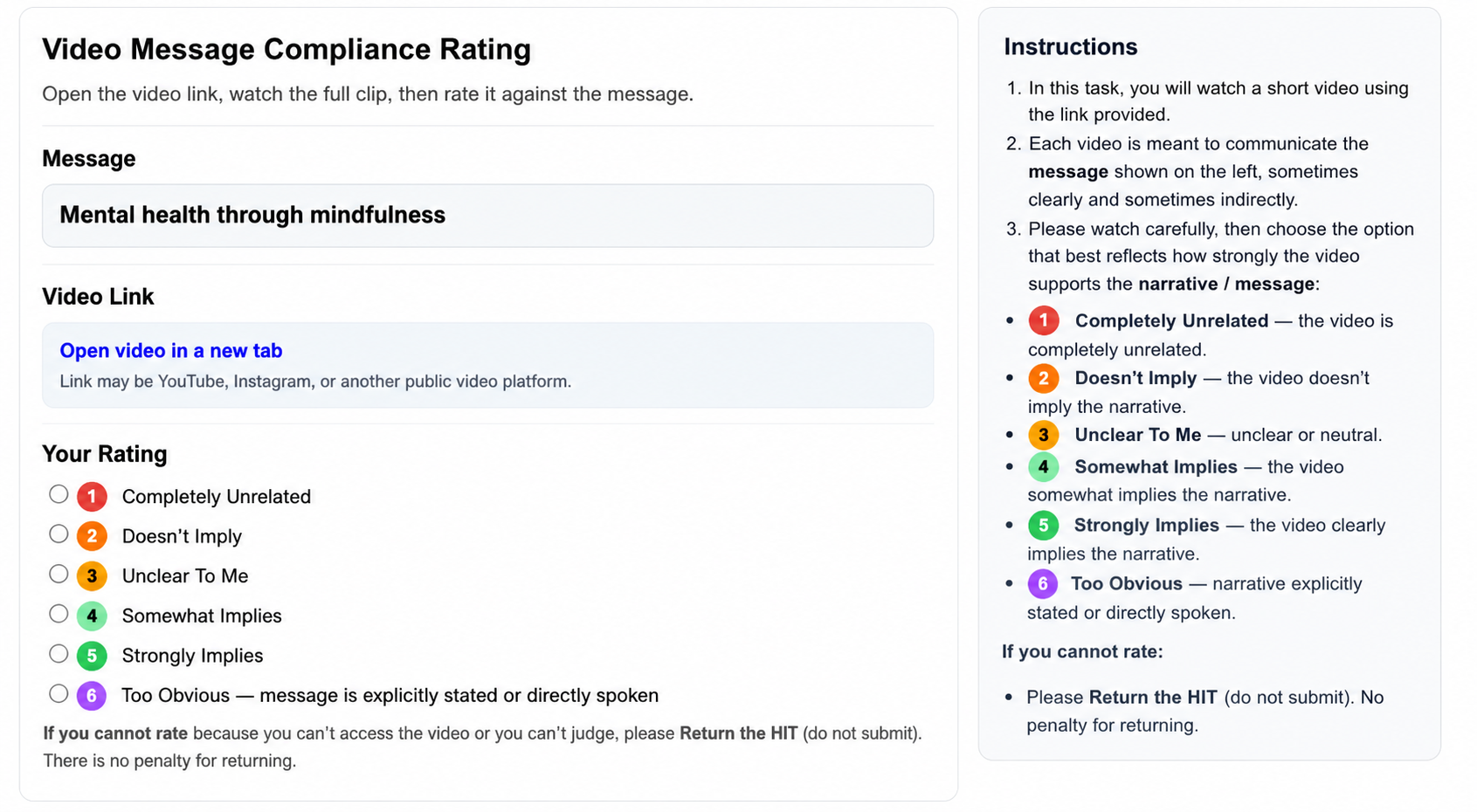}
    \captionsetup{font=small}
    \caption{Amazon Mechanical Turk Tagging Interface: A screenshot of a video message compliance rating task shown to a Mechanical Turk worker. The interface includes the main tagging panel, where the worker is asked to evaluate whether a video supports a message, along with a video link and rating options ranging from “Completely Unrelated” to “Too Obvious.” A separate instructions panel explains how to complete the task, what each rating level means, and what to do if the worker cannot access or judge the video.}
    \label{fig:amt-screenshot}
\end{figure}

After candidate clips are collected, we use human annotation to assess how well each video supports its corresponding message. Fig.~\ref{fig:amt-screenshot} shows the Amazon Mechanical Turk interface used for this validation step. Workers are shown the target message and a candidate video link, then asked to rate the degree of message support using an ordinal scale ranging from unrelated content to videos that express the message too directly. The accompanying instruction panel defines each rating level and provides guidance for cases where the video cannot be accessed or judged.

\subsection{
Crowdsourcing}
\begin{figure}[b]
    \centering
    \includegraphics[width=0.8\linewidth]{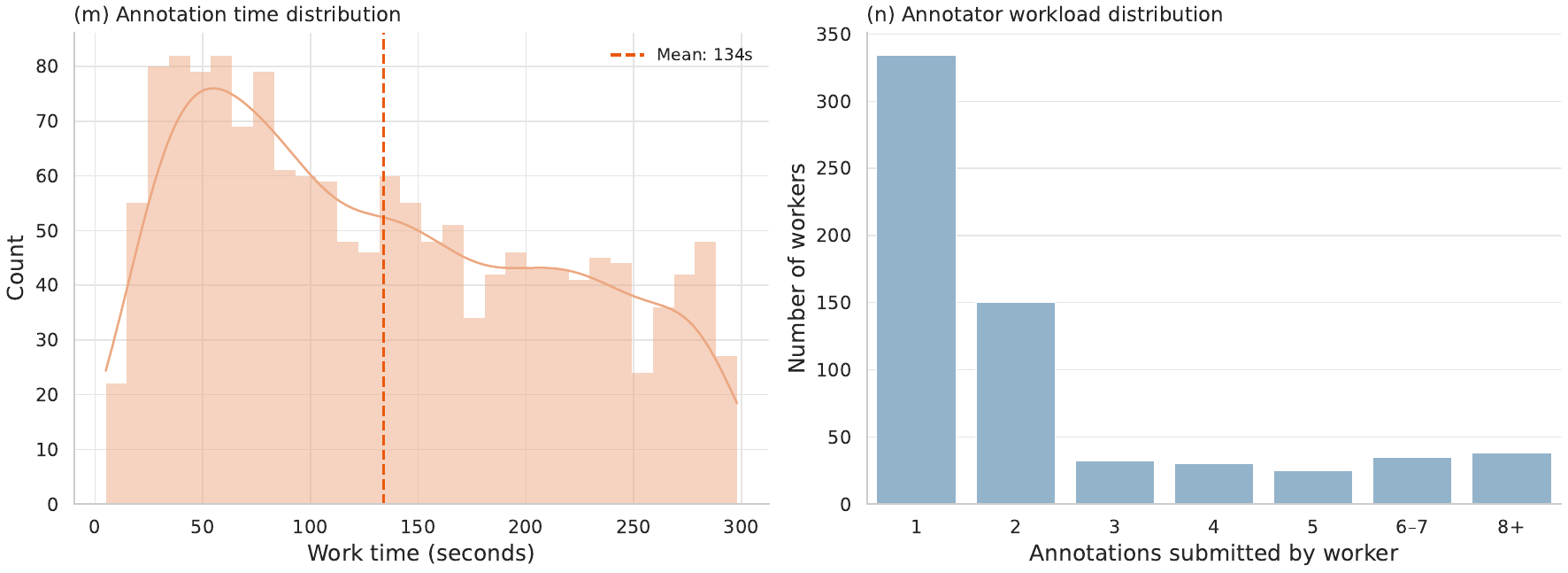}
    \captionsetup{font=small}
    \caption{Annotation effort and worker workload statistics. The figure shows the time required to tag videos with messages and the distribution of the number of annotations completed by each worker.}
    \label{fig:amt-time-workload}
\end{figure}

Fig.~\ref{fig:amt-time-workload} illustrates the time required to tag videos with messages, as well as the distribution of the number of annotations per worker. Workers were compensated through an auction bid.

\subsection{LLM Prompts}
\label{app:llm-prompts}
We detail the prompts used throughout the LLM-assisted construction pipeline. Starting from a target message, the pipeline first generates a short narrative scenario in which the message is expressed implicitly rather than directly; the prompts for this stage are shown in Figure~\ref{fig:prompt-story-generation}. The generated story can then be refined through an additional feedback-based prompting step, shown in Figure~\ref{fig:prompt-story-refinement}. Next, the story is converted into a concise YouTube search query using the prompts in Figure~\ref{fig:prompt-youtube-query-generation}, and the resulting query is checked for alignment with the original message using the validation prompts in Figure~\ref{fig:prompt-query-validation}. Finally, retrieved candidate clips are ranked according to their relevance to the target message using the prompts in Figure~\ref{fig:prompt-video-ranking}.

\section{Details on \ours}
Since collecting supervised training data for our task is difficult and costly, as it requires large-scale video clips paired with annotated implicit messages, we follow VidVec~\cite{vidvec2026} and adopt a lightweight \emph{text-only} optimization strategy. VidVec studies how pretrained multimodal large language models can be adapted into effective embedding models for video--text retrieval. Its key observation is that substantial video--text retrieval capabilities are already encoded in pretrained MLLMs. Based on this observation, VidVec avoids expensive video-based fine-tuning and instead applies a lightweight optimization stage using text-only supervision. Specifically, it aligns dense video captions with concise textual summaries, thereby adapting the embedding space to the target retrieval task while preserving the general semantic knowledge of the pretrained backbone.

We extend this idea to implicit message retrieval in a fully synthetic setting. Instead of optimizing on dense-caption--summary pairs associated with existing videos (as in VidVec), we synthesize dense-caption--message pairs. In each pair, the message is a short, interesting cultural statement, while the dense caption is a visual storyline describing a realistic video scenario that visually supports the message. This construction follows the same \emph{in-context} alignment principle used in VidVec: a rich contextual description of a clip is aligned with a compact semantic target. Optimizing on such pairs encourages the model to associate implicit messages with concrete narrative evidence. Importantly, the message is not explicitly stated in the caption; rather, it must be recoverable from the content. This makes the optimization closer to implicit message understanding than standard caption--text matching.

To increase the diversity of the generated data, we incorporate PersonaHub persona conditioning. PersonaHub~\cite{personahub} proposes large-scale persona-driven synthetic data generation, where heterogeneous personas are used to prompt language models to produce data reflecting diverse backgrounds, roles, viewpoints, and social contexts. This is particularly useful for our task, since cultural messages and implicit social meanings are strongly shaped by perspective, audience, and context. We therefore condition the generation process on different personas, allowing exposure to a broader range of cultural messages and visually grounded video scenarios. The persona is used only as a generation prior: the generated video caption must not include or depict the persona, and the inferred message must not depend on the viewer knowing who published the video. This ensures that the message is recoverable from the clip storyline itself rather than from external author identity.

We follow the same optimization configuration as VidVec~\cite{vidvec2026}, including the selected backbone and training hyperparameters. The synthetic optimization data were generated using the prompt shown in Fig.~\ref{fig:prompt-persona}.

\begin{figure}[b]
\centering
\begin{promptbox}
Generate a short human/cultural related message (5-7 words) that the following persona might express in a short video clip shared on a social network to a general audience.

The message should be related to cultural trends and debatable current notions. The message should be arguable. Keep the message clear.

Generate also the video clip dense captioning (explaining the visual alone). The video should convey the message, and should not include the persona itself or talk (speak) the message. The message should be crystal clear from the video storyline, and it should be supported by the video. The watcher won't know who published the video. Preferably realistic video.

The Persona:
{persona}

Note:

1. Output ONLY valid JSON with exactly two keys: "message" and "video_caption".
2. "message" should be the short 5-7 word arguable cultural message.
3. "video_caption" should be a dense description of the video visuals (no speech, no text overlays, no persona appearance).
4. Do not include any explanation, prefix, or extra text outside the JSON.
\end{promptbox}
\caption{Prompt used by \ours{} to generate PersonaHub-based training data.}
\label{fig:prompt-persona}
\end{figure}

\section{Implementation Details}
\label{sec:implementation}

Due to limited access to large-scale search and retrieval APIs for short-form social platforms such as TikTok, Instagram Reels, and X, we construct \bmname{} from public YouTube videos. YouTube provides a scalable and reproducible source of web video content while still containing many short, clips with dense communicative content. We therefore view \bmname{} as a YouTube-derived diagnostic benchmark whose motivation extends to short-form online video more broadly.

Candidate clips were retrieved using the YouTube API. We filtered retrieved videos by duration, retaining only clips shorter than three minutes. As described in Sec.~\ref{sec:data collection}, final inclusion was determined through human annotation rather than metadata alone.

For the message-first collection pipeline, we used GPT-5.1 for story generation, keyword generation, and judging. All prompts used for data construction are provided in~\ref{app:llm-prompts}.

Human annotation was conducted through Amazon Mechanical Turk (AMT). Annotators were restricted to workers located in the United States, at least 18 years old, and with a high task-acceptance rate. Each candidate clip was assigned to three independent annotators, who rated message relevance and flagged clips where the message was too explicit. The full annotation instructions, interface screenshot, and worker compensation details are provided in Sec.~\ref{sec:appendix-dataset-construction}.

For \ours{}, the synthetic optimization data were generated using Gemma 3~\cite{gemma}. The data consists of visual-storyline--message pairs designed to teach the model to associate implicit messages with narrative evidence. Training was performed on 4 NVIDIA B200 GPUs and completed in under 30 minutes. We follow VidVec~\cite{vidvec2026} for backbone selection and optimization settings.

For evaluation, each baseline was run using the input protocol and number of frames specified in its original paper or released code, with a maximum sampling rate of 1 FPS. This preserves each model's native operating setting while ensuring consistent temporal sampling. In the retrieval setting, models are evaluated using visual frames as input. For MLLM-based embedders that support multimodal textual context, we also evaluate a transcript-augmented setting, in which the complete transcription is prepended to the video frames. Transcripts are generated using openai/whisper-large-v3-turbo. In the MCQ evaluation, all models are evaluated using visual input only, except for Gemini models, which are run with native YouTube access. The specific GPT versions evaluated are gpt-5.4-Mini-2026-03-17 and gpt-5.4-2026-03-05.

\section{Limitations}
VidMsg has several limitations. Implicit message understanding is inherently subjective and may vary across viewers, cultures, languages, and contexts, despite our use of multiple AMT annotators and conservative filtering. The benchmark may be viewed as limited in scale, which may restrict the diversity of messages and video styles it covers. Yet, we believe it can be an initial step toward evaluating message-level video understanding. In addition, the LLM-assisted collection pipeline may introduce biases through generated scenarios, search queries, and metadata-based retrieval. Therefore, the annotations should be interpreted as consensus-based evaluation labels rather than exhaustive ground truth for all possible interpretations of each video.

\section{Societal Impact}
Message retrieval in videos can improve access to socially meaningful visual content, such as educational media, awareness campaigns, and short-form videos that convey implicit social or emotional messages. However, inferring latent messages is inherently subjective and may reflect cultural biases, annotator assumptions, or missing context, particularly in online shared videos. Such capabilities could also be misused for persuasive targeting, surveillance, or profiling. We therefore position \bmname primarily as a research benchmark and encourage future uses of message retrieval systems to consider fairness, privacy, cultural context, and transparency.

\section{Licenses}
The video clips are licensed under the Standard YouTube License. We release our final annotations under a CC-BY-4.0 license, together with the corresponding YouTube URLs.

\captionsetup[subfigure]{position=bottom}

\begin{figure}
\centering

\begin{subfigure}{0.95\textwidth}
\begin{promptbox}
You are a creative video storyteller specializing in short video content for public platforms.

Your task is to create a brief story concept that subtly incorporates a hidden narrative.

Requirements:
- Style: {style}
- Keep it very concise: 2-3 sentences maximum
- Focus on the core story idea that can be expanded later
- The hidden narrative should be woven naturally into the concept
- Suitable for short video platforms

Provide only a brief story concept, not a detailed outline.
\end{promptbox}
\caption{System prompt}
\end{subfigure}

\vspace{0.75em}

\begin{subfigure}{0.95\textwidth}
\begin{promptbox}
Create a brief story concept (2-3 sentences) that incorporates this hidden narrative:

Hidden Narrative: "{narrative}"

Provide only a concise story idea that:
1. Subtly includes the hidden narrative
2. Is suitable for short video content
3. Can be expanded into a full video later

Keep it short and focused on the core concept only.

Additional Context: {additional_context}
\end{promptbox}
\caption{User prompt}
\end{subfigure}

\caption{Prompts used for brief story generation.}
\label{fig:prompt-story-generation}
\end{figure}

\begin{figure}[]
\centering

\begin{subfigure}{0.95\textwidth}
\begin{promptbox}
You are a video story editor. Your task is to refine an existing story based on specific feedback while maintaining the core narrative and video suitability.
\end{promptbox}
\caption{System prompt}
\end{subfigure}

\vspace{0.75em}

\begin{subfigure}{0.95\textwidth}
\begin{promptbox}
Please refine this video story based on the following request:

Original Story:
{original_story}

Hidden Narrative: {narrative}

Refinement Request: {refinement_request}

Please provide the refined story that addresses the feedback while maintaining the hidden narrative and video format suitability.
\end{promptbox}
\caption{User prompt}
\end{subfigure}

\caption{Prompts used for story refinement.}
\label{fig:prompt-story-refinement}
\end{figure}

\begin{figure}
\centering

\begin{subfigure}{0.95\textwidth}
\begin{promptbox}
You are a YouTube search expert. Your task is to generate ONE optimized search query based on a story.

Instructions:
- Your goal is to create a search query (2-6 words) that helps find relevant YouTube videos based on the story content.
- Focus on the concrete elements, actions, and scenarios described in the story.
- Favor phrases that reflect change, action, or emotional stakes (e.g., quitting, overcoming, learning, rediscovering), rather than static descriptors.
- Avoid generic terms. Use specific, searchable phrases that reflect the unique aspects of the story.
- Do not include quotes, formatting, or additional commentary.

Examples:
Story: A burnt-out office worker rediscovers their love for music and begins performing at open mics.
-> "office worker musician transition"

Return ONLY the search query. Nothing else.
\end{promptbox}
\caption{System prompt}
\end{subfigure}

\vspace{0.75em}

\begin{subfigure}{0.95\textwidth}
\begin{promptbox}
Story: "{story}"

Based on the story, generate ONE YouTube search query that would surface videos reflecting the core scenario or theme.

NOTE: The following queries were already tried and rejected. Generate a DIFFERENT query:
"{query1}", "{query2}", ...
\end{promptbox}
\caption{User prompt}
\end{subfigure}

\caption{Prompts used for YouTube search-query generation.}
\label{fig:prompt-youtube-query-generation}
\end{figure}

\begin{figure}
\centering

\begin{subfigure}{0.95\textwidth}
\begin{promptbox}
You are a content alignment validator. Your task is to determine if a YouTube search query adequately represents a given narrative theme.

Evaluate whether the search query:
1. Captures the core concept of the narrative
2. Would find videos that align with the narrative's message
3. Is specific enough to be useful but not so narrow it misses the point

Respond with ONLY a JSON object in this exact format:
{"valid": true, "reasoning": "Brief explanation"}
or
{"valid": false, "reasoning": "Brief explanation of why it doesn't represent the narrative"}

Be reasonably lenient - the query doesn't need to be perfect, just representative of the narrative theme.
\end{promptbox}
\caption{System prompt}
\end{subfigure}

\vspace{0.75em}

\begin{subfigure}{0.95\textwidth}
\begin{promptbox}
Narrative: "{narrative}"

Search Query: "{search_query}"

Does this search query adequately represent the narrative theme? Respond with JSON only.
\end{promptbox}
\caption{User prompt}
\end{subfigure}

\caption{Prompts used for search-query validation.}
\label{fig:prompt-query-validation}
\end{figure}

\begin{figure}
\centering

\begin{subfigure}{0.95\textwidth}
\begin{promptbox}
You are a video content analyst specializing in narrative alignment. Your task is to analyze YouTube videos and rank them based on their relevance to a given narrative theme.

For each video, you will analyze:
1. Title - How well does it align with the narrative?
2. Description - Does the content description support the narrative?
3. Uploader - Is the channel type relevant to the narrative?
4. Duration - Is the video length appropriate for the narrative content?
5. View count - Does popularity indicate relevance?

You must provide a JSON response with the following structure:
{
  "rankings": [
    {
      "video_id": "video_id_here",
      "relevance_score": 8.5,
      "relevance_reasoning": "Detailed explanation of why this video is relevant to the narrative"
    }
  ]
}

Scoring criteria:
- 9-10: Perfectly aligned with narrative, highly relevant content
- 7-8: Strong alignment, clearly relevant
- 5-6: Moderate alignment, somewhat relevant
- 3-4: Weak alignment, tangentially relevant
- 1-2: Poor alignment, barely relevant

Focus on narrative alignment over video quality or popularity. Be thorough in your reasoning.
\end{promptbox}
\caption{System prompt}
\end{subfigure}

\vspace{0.75em}

\begin{subfigure}{0.95\textwidth}
\begin{promptbox}
Narrative: "{narrative}"

Videos to analyze and rank:
{video_data_as_json}

Please analyze each video's relevance to the narrative and provide rankings in the specified JSON format. Consider how well each video's content, as indicated by its title, description, and context, aligns with the narrative theme.

Focus on:
1. Thematic alignment with the narrative
2. Content relevance based on available metadata
3. Appropriateness of the video format for the narrative
4. Potential storytelling value

Provide your response as valid JSON only.
\end{promptbox}
\caption{User prompt}
\end{subfigure}

\caption{Prompts used for video ranking.}
\label{fig:prompt-video-ranking}
\end{figure}

\end{document}